\documentclass[10pt,twocolumn,letterpaper]{article}

\usepackage[pagenumbers]{cvpr} 

\usepackage{tikz}
\usetikzlibrary{shapes.geometric, arrows}

\tikzstyle{startstop} = [rectangle, rounded corners, 
minimum width=3cm, 
minimum height=1cm,
text centered, 
draw=black, 
fill=red!30]

\tikzstyle{io} = [rectangle, rounded corners, 
minimum width=2.5cm, 
minimum height=1cm,
text centered, 
draw=black, 
fill=blue!30]

\tikzstyle{process} = [rectangle, rounded corners, 
minimum width=2.5cm, 
minimum height=1cm,
text centered, 
draw=black, 
fill=orange!30]

\tikzstyle{decision} = [diamond, 
minimum width=2.5cm, 
minimum height=1cm, 
text centered, 
draw=black, 
fill=green!30]
\tikzstyle{arrow} = [thick,->,>=stealth]

\definecolor{cvprblue}{rgb}{0.21,0.49,0.74}
\usepackage[pagebackref,breaklinks,colorlinks,allcolors=cvprblue]{hyperref}

\def\paperID{11938} 
\def\confName{CVPR}
\def\confYear{2025}

\title{A Modular Pipeline for 3D Object Tracking Using RGB Cameras}

\author{Lars Bredereke\\
University Bremen\\
Cognitive Systems Lab\\
{\tt\small labr@uni-bremen.de}
\and
Yale Hartmann\\
University Bremen\\
Cognitive Systems Lab\\
{\tt\small yale.hartmann@uni-bremen.de}
\and
Tanja Schultz\\
University Bremen\\
Cognitive Systems Lab\\
{\tt\small tanja.schultz@uni-bremen.de}
}

\begin{document}
\maketitle
\begin{abstract}
Object tracking is a key challenge of computer vision with various applications that all require different architectures. Most tracking systems have limitations such as constraining all movement to a 2D plane and they often track only one object. In this paper, we present a new modular pipeline that calculates 3D trajectories of multiple objects. It is adaptable to various settings where multiple time-synced and stationary cameras record moving objects, using off the shelf webcams.
Our pipeline was tested on the Table Setting Dataset, where participants are recorded with various sensors as they set a table with tableware objects. We need to track these manipulated objects, using 6 rgb webcams.
Challenges include: Detecting small objects in $9.874.699$ camera frames, determining camera poses, discriminating between nearby and overlapping objects, temporary occlusions, and finally calculating a 3D trajectory using the right subset of an average of $11.12.456$ pixel coordinates per \mbox{3-minute} trial. We implement a robust pipeline that results in accurate trajectories with covariance of x,y,z-position as a confidence metric. It deals dynamically with appearing and disappearing objects, instantiating new Extended Kalman Filters. It scales to hundreds of table-setting trials with very little human annotation input, even with the camera poses of each trial unknown. The code is available at \url{https://github.com/LarsBredereke/object_tracking}.
\end{abstract}    
\section{Introduction}
\label{sec:intro}

Multiple Object Tracking (MOT) and Multi-Camera Multiple Object Tracking (MCMOT) are one of the most complex forms of object tracking. In our case, the information of 6 overlapping cameras' recordings needs to be processed to calculate three-dimensional trajectories. Our implementation was tested on the Table Setting Dataset (TSD), but the modular nature of the pipeline makes it trivial to change the scenario to any other space recorded by time-synced, overlapping, and stationary cameras, where every tracked object should be visible to multiple cameras for most of the time. The pipeline can be adapted to a wide range of trackable objects, as YOLO-networks can detect a wide range of objects.

\begin{figure*}
  \centering
  \includegraphics[width=1\textwidth]{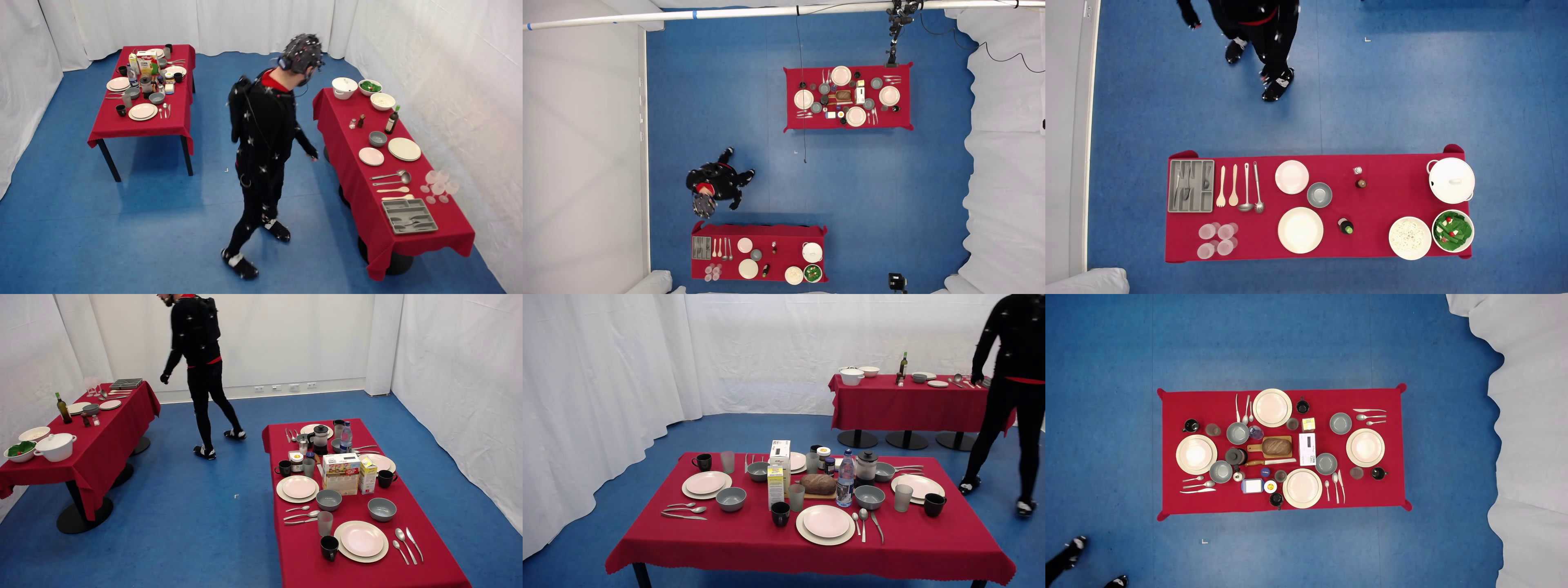}
  \caption{The TSD cameras used for this paper are referred to as back, ceiling, counter-top, front, table-side and table-top.}
  \label{fig:all_cameras}
\end{figure*}

The Table Setting Dataset provides multi-modal recordings for the activity of setting a table with tableware. To analyze table setting behavior, the human test subject is fitted with a motion capture suit, EEG, EMG, eyetracking and more. However, the manipulation of tableware objects is only recorded in video form. Therefore, actions performed by participants were previously only annotated manually, restricting the analysis of human behavior with machine learning. These manual annotations include timestamps, the hand used and the class of tableware object used. It does, however, not contain spacial information about where the object has been picked up and its trajectory. Extracting this spacial information of tableware objects from the provided webcam recordings shown in \Cref{fig:all_cameras} is the subject of this paper.

Methods used include object detection using a YOLO-network, a gradient descent to optimize the camera's position and orientation, and Kalman Filters to find the most likely trajectory of each object given the observations.


\section{Related Research}
\label{sec:related_research}

In computer vision, object tracking is a classic challenge with many applications for extracting trajectories from video streams. However, most object trackers focus on tracking in some two-dimensional space like the image plane or the ground plane, with comparatively few papers focusing on the full three degrees of freedom in movement.

Alex Bewley et al~\cite{Bewley_2016} use a pipeline with a similar concept of a modular pipeline, where image detection algorithms predict bounding boxes, which are then processed with Kalman Filters. While our pipeline is built for post processing footage, this paper shows that such a pipeline can be easily developed into a real-time application. However, their approach only considers two-dimensional tracking.

Cheng-Che Cheng et al~\cite{cheng2023restreconfigurablespatialtemporalgraph} track multiple people on a 2D plane while dealing with temporary occlusions. Conceptually, they divide tracking into spatial and temporal association, allowing them to better deal with temporary occlusions. Similarly, Duy M. H. Nguyen et al~\cite{nguyen2022lmgpliftedmulticutmeets} track bounding boxes on a two-dimensional plane in three-dimensional space. Tracklets of individual cameras are then combined into trajectories.

Moving to Multi-Camera Multi Object Tracking (\mbox{MCMOT}), Shuxiao Ding et al~\cite{ding2024adatrackendtoendmulticamera3d} utilize cameras of a moving car to track objects in the street. However, they do not use explicit knowledge about camera poses to triangulate objects, as they use neural networks for this task.  

Moving back to single object tracking, Jingtong Li et al~\cite{li2020reconstruction3dflighttrajectories} track a drone from multiple cameras. They use 2D-tracklets of drones in image-coordinates to generate a 3D-trajectory, without the cameras being in sync. They also deal with cameras being in an unknown position. Noise in the trajectory is reduced by penalizing acceleration in 3D space. This paper does not fit our need for a fixed world-coordinate system, time-syncing and discriminating between identical looking objects.

Jérôme Berclaz et al~\cite{5708151} have a similar goal to us, as they track multiple objects in 3D space. However, they discretize 3D space and assume that two objects can not share the same position, which does not hold true for our use case, where the trajectories of objects can converge or diverge, as objects like plates are stackable.

Finally, Fatih Cagatay Akyon et al~\cite{Akyon_2022} detect small objects in large images, which is a challenge that we also face with small tableware objects being recorded from across the room. However, the YOLO network we use can already detect small objects out of the box, even when objects are only a few pixels wide.

\section{Data}
\label{sec:data}
The Dataset used as input for our pipeline is the Everyday Activity in Science and Engineering (EASE) Table Setting Dataset (TSD)~\cite{mason_iros_2018,mason_iros2020}, which records humans setting a table. Dozens of sessions are provided, each containing around 6 trials. Each trial is a recording of a subject who is setting an initially empty table with tableware objects from a nearby counter. Depending on the trial instructions, the table is to be set for a varying number of guests with the type of meal also changing from trial to trial.

\begin{figure}[h]
    \centering
    \includegraphics[width=1\linewidth]{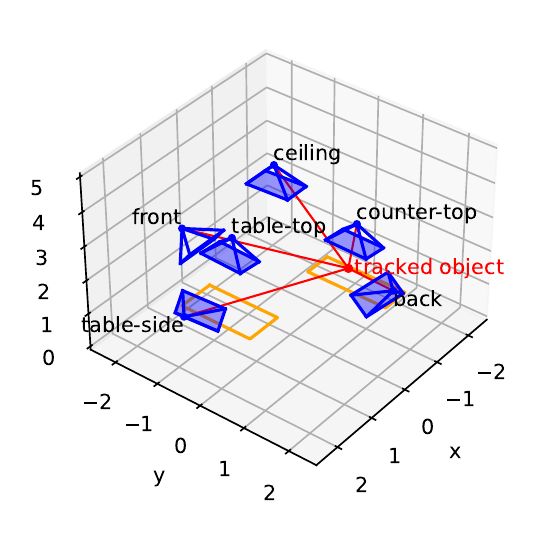}
    \caption{Camera and table poses with a tracked object} \label{fig:cam_params}
\end{figure}

One of the sensor modalities provided in the TSD are 6 cameras. They are stationary and record 720p at 30fps, with time-syncing provided by the Labstreaminglayer~\cite{lsl}. There currently exist dozens of sessions with 6 trials lasting about 3 minutes each. An example for recorded images can be seen in Figure \ref{fig:all_cameras}, with the camera poses being shown in Figure \ref{fig:cam_params}. The cameras are movable and can slightly change their position and orientation between trials. For some trials, individual cameras are not operational. In such cases, our pipeline works on only 5 cameras.

\section{Proposed Pipeline}
\label{sec:pipeline}
\begin{figure}
    \centering
    \begin{tikzpicture}[node distance=2cm]
    
        \node (start) [process, align=left] {\cref{sec:data}: video frames\\of 6 cameras at 30fps};
        \node (yolo) [process, below of=start, yshift=0.7cm, align=left] {\cref{sec:yolo}: YOLO 2D\\object detection};
        \node (annot) [process, right of=start, xshift=3.1cm, align=left] {\cref{sec:annot}: annotate \\ table corners \& orgin};
        \node (params) [process, below of=annot, yshift=0.7cm, align=left] {\cref{sec:optimize_parameters}: calculate\\exact camera pose};
        \node (ekf) [process, below of=yolo, xshift=2.6cm, yshift=0.7cm, align=left] {\cref{sec:find3d,sec:track3d}: calculate trajectories\\using heuristics and Kalman Filters};
    
        \draw [arrow] (start) -- (yolo);
        \draw [arrow, transform canvas={yshift=0.4cm}] (start) -- node[anchor=north, align=center] {1 image per \\ trial $\times$ cam} (annot);
        \draw [arrow] (annot) -- (params);
        \draw [arrow] (params) -- (ekf);
        \draw [arrow] (yolo) -- (ekf);

    \end{tikzpicture}
    \caption{Stages of the object tracking pipeline with references to their Sections} \label{fig:pipeline}
\end{figure}
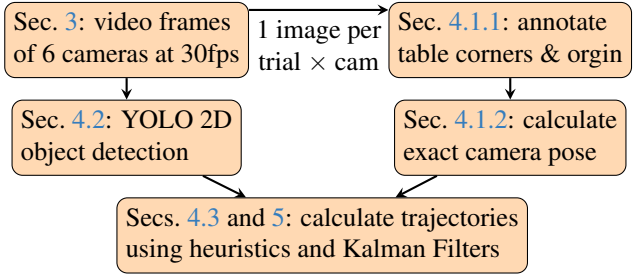

Extracting 3D trajectories from multiple rgb cameras takes many steps which are shown in \Cref{fig:pipeline}. This modular approach allows us to develop and optimize every step of the pipeline individually. For example, we can optimize our heuristics for \Cref{sec:find3d} without having to rerun the YOLO network on millions of frames. The following subsections are an abstracted description of the code, which is provided as supplementary material.

\subsection{Camera Parameters}
\label{sec:cam_par}
For our finished pipeline, we want to make a connection between pixel coordinates in a camera image and positions in 3D space. This requires us to know the exact position, orientation, focal length and radial distortion parameters of the cameras. The focal length and distortion are constant but position and orientation vary between the 320 trials. Therefore, we designed a calibration system that requires minimal user input and determines the exact position and orientation of the 6 cameras, only using images similar to those of \Cref{fig:all_cameras}. Note: Before calibrating the camera pose for each trial, the focal length and radial distortion parameters $k1$ and $k2$ are calibrated once as they are constant for all cameras and trials. The 'project' function mapping a point in 3D world-coordinates to image coordinates is largely identical to the implementation of OpenCV \cite{opencv_library}. However, we implemented it ourselves in order to be able to derive it using Tensorflow \cite{tensorflow2015-whitepaper}, as will be needed in \Cref{sec:optimize_parameters}.

\subsubsection{Annotation}
\label{sec:annot}
\begin{figure}[h]
    \centering
    \includegraphics[width=1\linewidth]{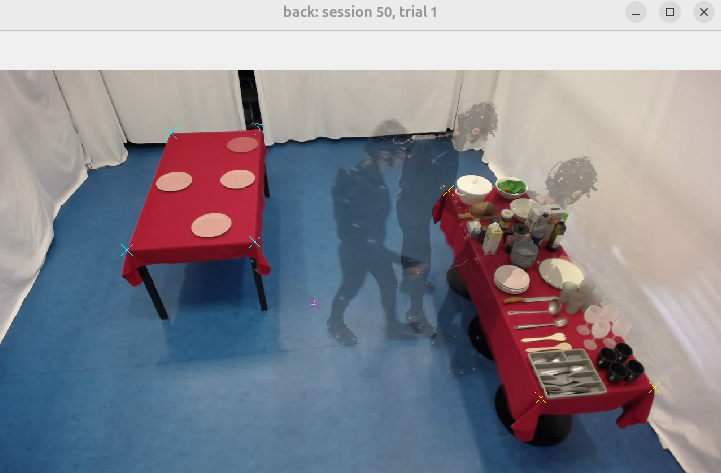}
    \caption{Screenshot of annotation tool used for labeling table(cyan), counter(yellow) and world origin(magenta) pixel positions} \label{fig:annotatetool}
\end{figure}

For each trial and camera, we manually annotate the pixel positions of all visible table corners as well as the world origin marker, if visible. We wrote a small GUI shown in \Cref{fig:annotatetool} that lets us click on the points to annotate and lets us skip re-annotating if the camera was not moved between trials. We also averaged over multiple still images to avoid occlusions of points to be annotated. Using this GUI, hundreds of trials can be annotated within a few hours.

\subsubsection{Optimizing Camera Poses}
\label{sec:optimize_parameters}
As the cameras can be slightly moved between trials, we need to solve camera positions and orientations. This builds on the principle that we can calculate where features annotated in \Cref{sec:annot} should appear in the image, given a camera pose and the feature's position in the world.

Using an initial guess shown in \Cref{tab:parameters}, we can calculate the expected pixel position of each feature. Subtracting that from the annotated pixel position gives us an error. Using Tensorflow, we calculate the gradient of that error to improve on our initial guess. After a few dozen iterations, we minimized the error and have calculated our parameters for a trial, as shown in \Cref{tab:optimized_parameters}. Note: The table's offsets to their expected positions are also optimized parameters because they can slightly move between trials. Also note: All parameters of a trial are optimized together with a single error function, allowing us to \eg calculate the pose of the counter-top camera that can only see the counter. (The exact position of which is also subject to optimization) Since we also annotate the world origin for some cameras, we can include it in our error function. This means our optimized camera poses are in the same world coordinate system that is also used for the motion capture suit of the test subject.

\begin{figure*}
    \centering
    \begin{subfigure}[b]{0.43\textwidth}
        \centering
        \begin{tabular}{ | c | c c c c c c |}
            \hline
            camera & x & y & z & pan & tilt & roll\\
            \hline
            table-side & 3 & -1 & 2 & $\frac{-\pi}{2}$& $\frac{-\pi}{7}$ & 0 \\
            table-top & 1 & -1 & 3 & $\frac{\pi}{2}$ & $\frac{-\pi}{2}$ & 0 \\
            back & 0 & 2.5 & 3 & $\pi$ & $\frac{-\pi}{4}$ & 0\\
            counter-top & -1.5 & 0.5 & 3 & $\frac{\pi}{2}$ & $\frac{-\pi}{2}$ & 0 \\
            ceiling & 0 & 0 & 5 & 0 & $\frac{-\pi}{2}$ & $\frac{\pi}{2}$ \\
            front & 0.5 & -2.5 & 3 & $\frac{-\pi}{12}$ & $\frac{-\pi}{4}$ & 0 \\
            \hline
            \hline
            table & \multicolumn{2}{c}{$x$ offset} & \multicolumn{2}{c}{$y$ offset} && \\
            \hline
            table & \multicolumn{2}{c}{0} & \multicolumn{2}{c}{0} && \\
            counter & \multicolumn{2}{c}{0} & \multicolumn{2}{c}{0} && \\
            \hline
        \end{tabular}
        \caption{Initial guess in meters and rad} \label{tab:parameters}
    \end{subfigure}
    \begin{subfigure}[b]{0.52\textwidth}
        \centering
        \begin{tabular}{ | c | c c c c c c |}
            \hline
            camera & x & y & z & pan & tilt & roll \\
            \hline
            table-side & 2.16 & -0.74 & 2.11 & -1.51 & -0.48 & 0.03 \\
            table-top & 0.70 & -0.80 & 2.79 & 2.17 & -1.50 & -0.58 \\
            back & -0.04 & 2.34 & 2.62 & 3.48 & -0.65 & -0.05 \\
            counter-top & -1.37 & 0.61 & 2.52 & 1.41 & -1.51 & 0.17 \\
            ceiling & 0.08 & -0.21 & 4.87 & -97.33 & -1.65 & 98.90 \\
            front & 0.33 & -2.35 & 1.97 & -0.11 & -0.62 & 0.06 \\
            \hline
            \hline
            table & \multicolumn{2}{c}{$x$ offset} & \multicolumn{2}{c}{$y$ offset} && \\
            \hline
            table & \multicolumn{2}{c}{-0.01} & \multicolumn{2}{c}{0.06} && \\
            counter & \multicolumn{2}{c}{-0.04} & \multicolumn{2}{c}{-0.02} && \\
            \hline
        \end{tabular}
        \caption{Optimized parameters for session 22 trial 1} \label{tab:optimized_parameters}
    \end{subfigure}
    \caption{Example of camera and table parameters before and after optimization for a trial}
\end{figure*}



\subsection{Object Detection in 2D Images}
\label{sec:yolo}
We want to be able to detect all tableware items that can be manipulated during the trials. They were not fitted with any sensors or motion capture markers and will have to be detected in the webcams' images. The TSD includes the following tableware objects that we want to detect: \texttt{bowl--salad}, \texttt{bowl--cooker}, \texttt{plate--pasta}, \texttt{bread}, \texttt{butter}, \texttt{jam}, \texttt{nutella}, \texttt{salt}, \texttt{shaker--pepper}, \texttt{sugar}, \texttt{cereal}, \texttt{milk}, \texttt{coffee}, \texttt{wine--bottle}, \texttt{water}, \texttt{bowl--cereal}, \texttt{plate}, \texttt{knife-bread}, \texttt{(utensil, pasta)}, \texttt{ladle}, \texttt{(utensil-spoon, salad)}, \texttt{(utensil-fork, salad)}, \texttt{teaspoon}, \texttt{tablespoon}, \texttt{fork}, \texttt{knife}, \texttt{glass-water}, \texttt{glass--wine}, \texttt{cup-coffee}.

\begin{figure}
    \centering
    \begin{subfigure}[b]{0.98\linewidth}
        \centering
        \includegraphics[clip, trim=9.3cm 0cm 0cm 0cm, width=1\linewidth]{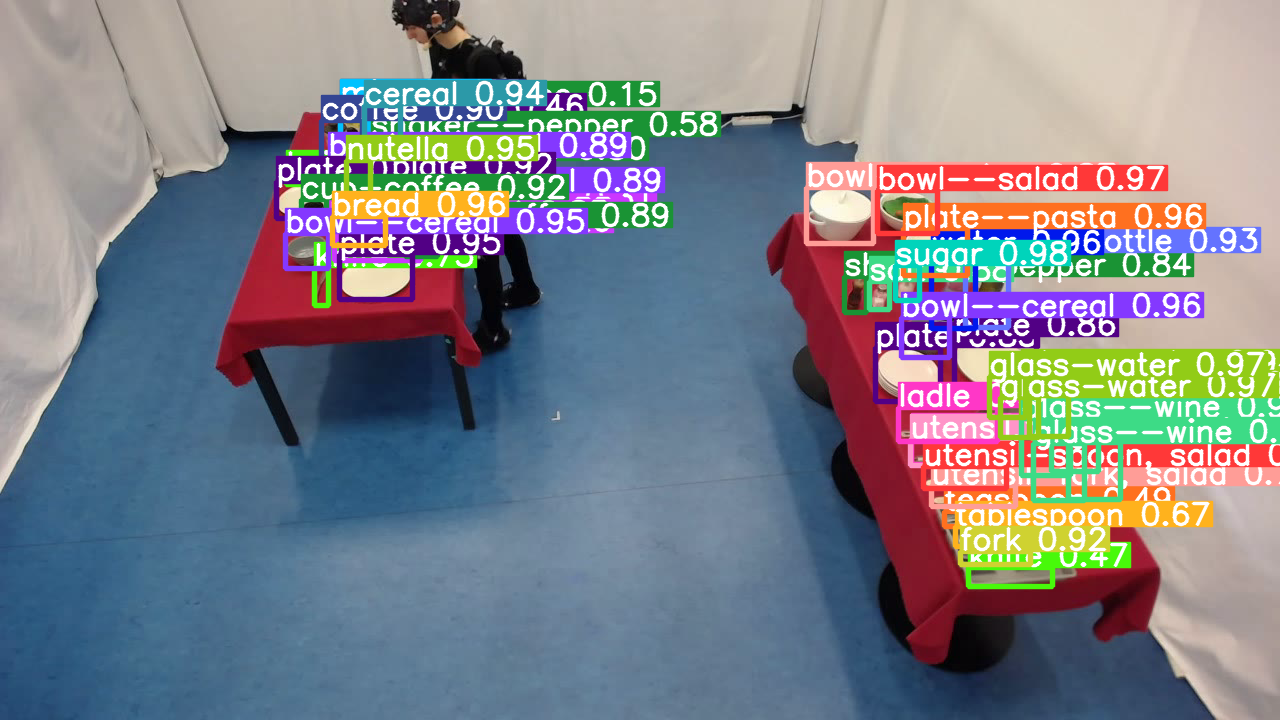} 
        \caption{Prediction of YOLO network on test data} \label{fig:yolo_detections}
    \end{subfigure}

    \begin{subfigure}[b]{0.98\linewidth}
        \centering
        \includegraphics[clip, trim=2.9cm 0cm 2.9cm 0cm, width=1\linewidth]{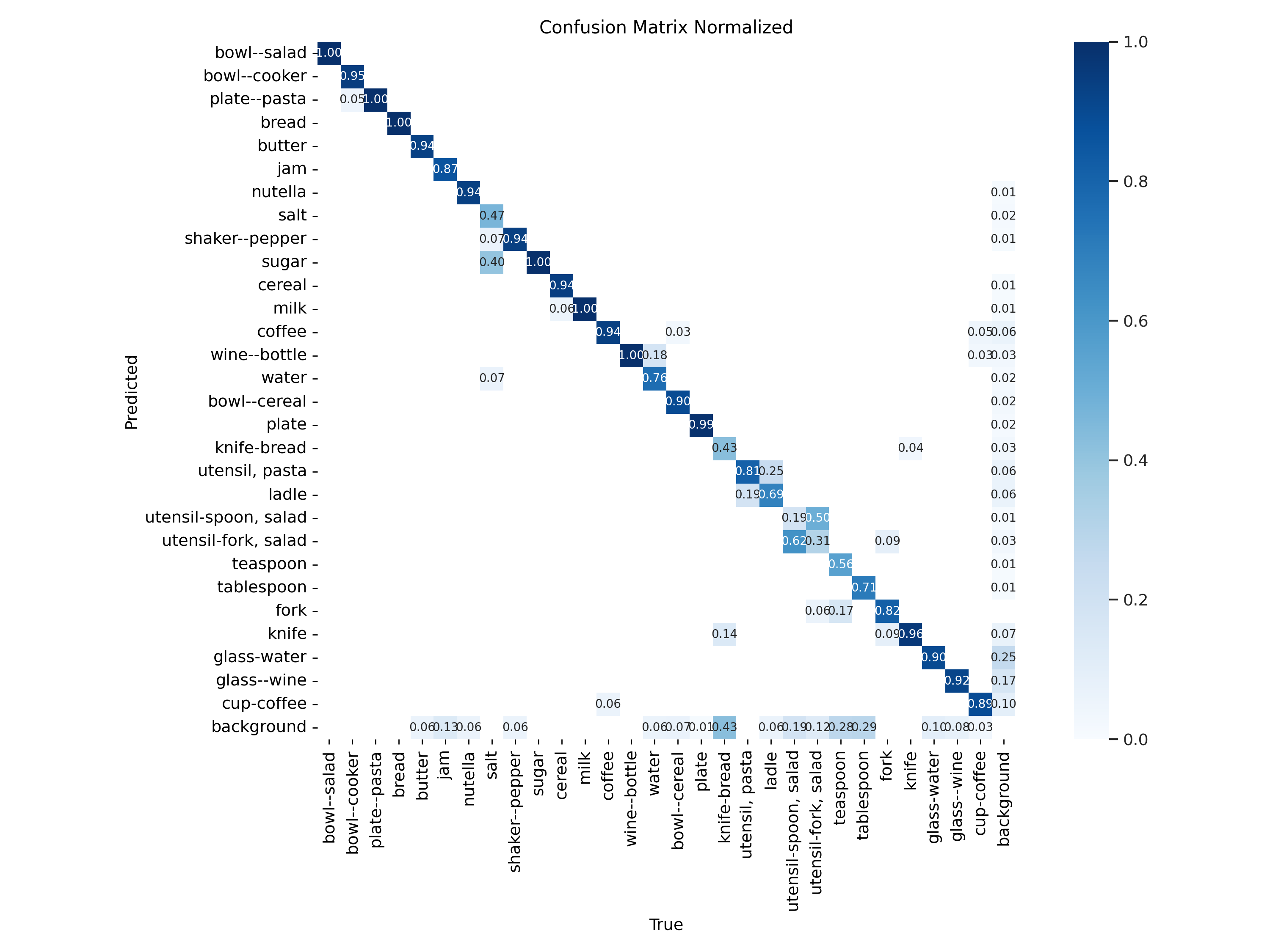} 
        \caption{Confusion Matrix of YOLOv8m model on test data} \label{fig:finalyolo}
    \end{subfigure}
    \caption{Performance of YOLO network on test data}
    \label{fig:yolo_fig}
\end{figure}

To automate the detection of tableware objects in images, we use a YOLOv8~\cite{yolov8_ultralytics} model, specifically the YOLOv8m~\footnote{\url{https://github.com/ultralytics/assets/releases/download/v0.0.0/yolov8m.pt}} model. For its training, we sampled 474 images from the 6 webcams of the TSD. We then manually annotated them with 16,451 bounding boxes, each labeled with a class of tableware. \Cref{fig:yolo_detections} shows the network's predictions on an image of the test dataset, which is from a session not present in the training dataset. This visualizes the challenge of detecting many small objects close to each other. The confusion matrix on the test dataset can be seen in~\Cref{fig:finalyolo}. The f1-score on the test dataset is $0.83$, using a confidence threshold of $0.5$. Given this confusion matrix, we decided to combine 'sugar' and 'salt' to 'salt/sugar' as well as to combine 'utensil-spoon, salad' and 'utensil-fork, salad' to 'utensil, salad', as they are hard to discriminate even for humans. Propagating all $9.874.699$ images through the YOLO network took 6 days on a NVIDIA GeForce GTX 1080 Ti with 12 Gigabytes of VRAM. A total of $355.986.035$ object detections were made.

\subsection{Detecting New Objects in 3D Space}
\label{sec:find3d}
\begin{figure}
    \centering
    \begin{subfigure}[b]{0.48\linewidth}
        \centering
        \includegraphics[clip, trim=0cm 3cm 0cm 2.5cm, width=1\linewidth]{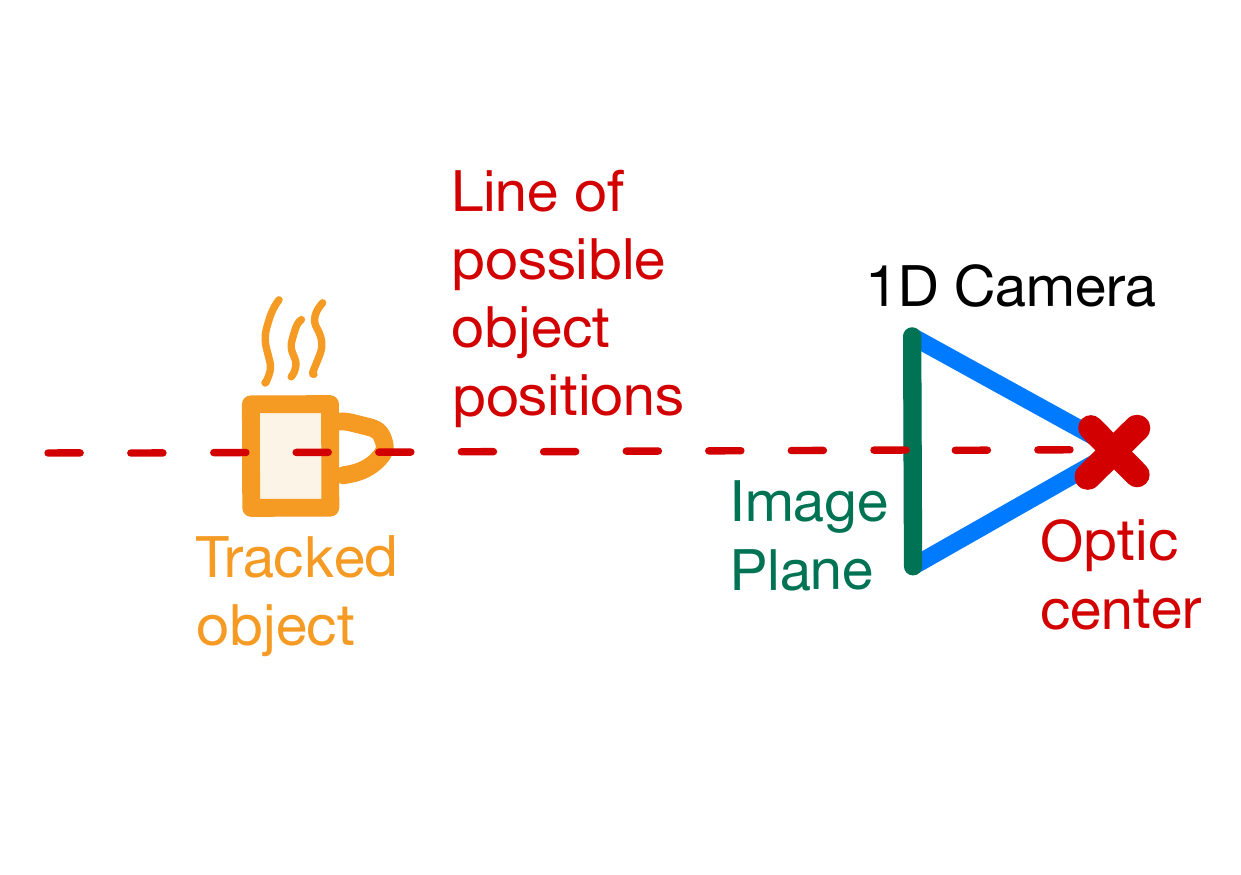}
        \caption{2D Scenario where we know the camera's position and want to locate a coffee cup visible in the camera's image}
        \label{fig:approach1}
    \end{subfigure}
    \hfill
    \begin{subfigure}[b]{0.48\linewidth}
        \centering
        \includegraphics[clip, trim=0cm 1cm 0cm 1cm, width=1\linewidth]{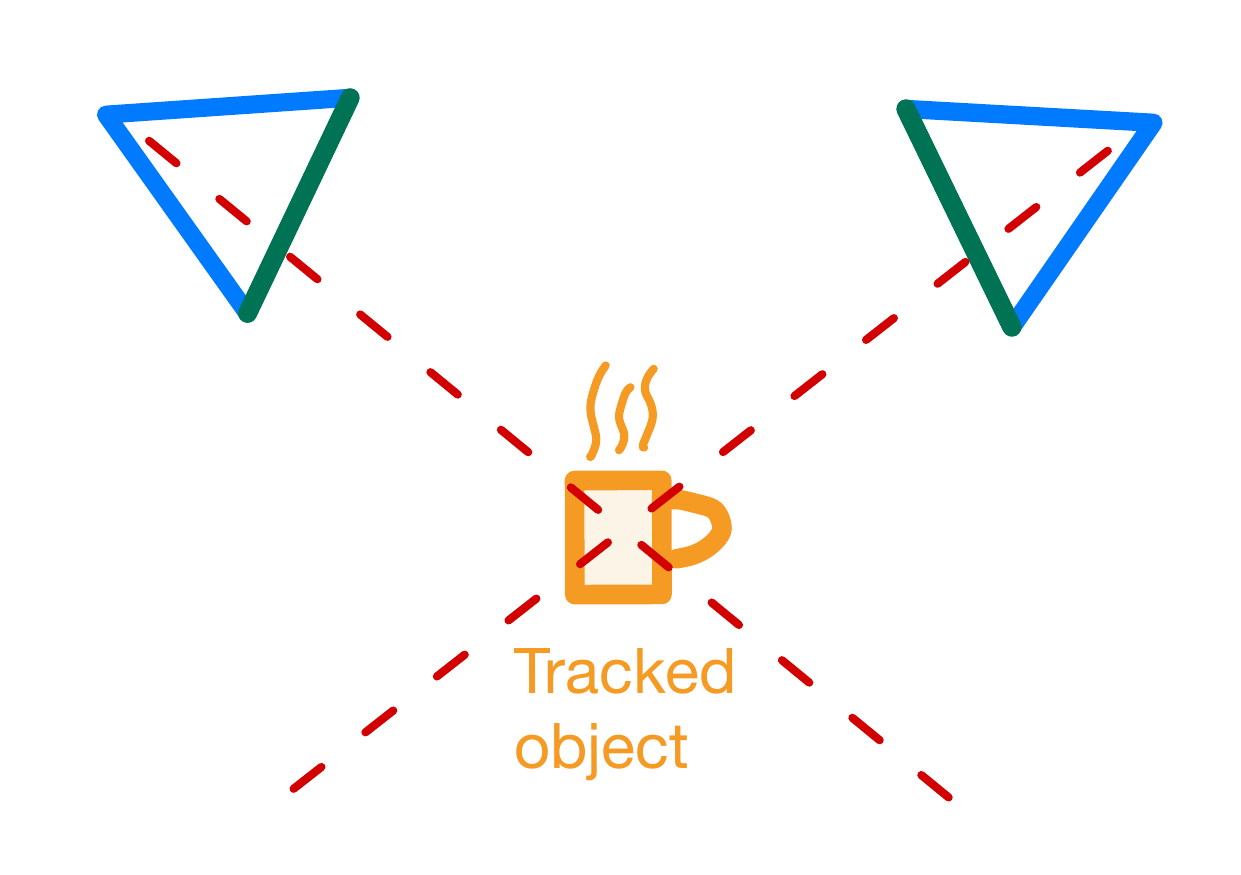}
        \caption{Extended 2D Scenario with two cameras}
        \label{fig:approach2}
    \end{subfigure}
       \caption{Camera Illustration in a 2D world}
       \label{fig:approach}
\end{figure}

We now have over 100 YOLO-detections per timestep spread over 6 images. As \Cref{fig:approach} illustrates for a 2D scenario, we can deduce an object's position by it's appearance in multiple camera's images. To find new objects, for each timestep, we take all YOLO-detections that are not already associated with a tracklet and perform the following steps:

\begin{itemize}
    \item Interpret every YOLO-detection's pixel coordinates as a line in three-dimensional space starting from the camera's position. The direction of the line is given by the pixel coordinates and the camera's orientation and focal length.
    \item Find intersections of lines with the same tableware class. This is has some challenges:
    \begin{itemize}
        \item The lines come close to each other but do not actually intersect. So what we actually do is walk along the lines and check if other lines of the same tableware class come within a threshold distance.
        \item If the tableware class only exists once in the world(\eg the box of cereal), the intersection of two lines will correspond to an object's position. But if the class has multiple objects, this might lead to false detections. (Imagine two coffee cups in Figure \ref{fig:approach2}, leading to line intersections that do not correspond to an object.) So for tableware classes that appear multiple times, we require detection lines from at least 3 different cameras. Note: This constraint only applies to initial detection. Tracking a previously detected object can be done with fewer cameras.
        \item If two lines are nearly parallel, then we can not derive the point of intersection accurately in all three dimensions. Therefore, a pair of nearly parallel lines counts as one line when evaluating the previous 2-camera or 3-camera constraint.
    \end{itemize}
    \item Delete intersections that are too close to existing tracklets of the same tableware class. If we initiated a tracklet here, then it would probably converge with the existing one, costing us runtime to track the same object multiple times. On the other hand, if the YOLO detections and camera poses are really accurate, we do want to be able to track close objects of the same tableware class. So once every 100 frames we drastically reduce the minimum distance required. We later delete tracklets if they converge with the position of another object shortly after initial detection. Note: Allowing for convergence of objects is necessary when \eg two plates are stacked.
    \item Instantiate a new tracklet for every remaining intersection, that was not excluded by one of the previous steps. This will happen at the start of the trial, but also when an object reappears from long motion blur or \eg plates appear from a stack of plates.
\end{itemize}

\subsection{Tracking detected objects in 3D space over time}
After initial detection of an object in 3D space, we need to map subsequent 2D YOLO-detections to it and update its 3D position. This done using Extended Kalman Filters.

\subsubsection{Extended Kalman Filters}
Each tracklet is modeled by a Kalman Filter. In many scenarios, a Kalman Filter (KF) is the mathematically ideal tool to predict a belief state (position of our object in world coordinates) given the measurements up to the current point in time. It also gives us a covariance matrix which lets us know how confident we can be with the KF's belief state. In our case, the camera function mapping the state space (world coordinates) to the measurement space (pixel coordinates) is non-linear. KFs however require a linear function. An Extended Kalman Filter (EKF) can linearize the camera function by taking its derivative at our current belief state. This works well for us because the camera function is close to linear in the relevant value ranges.
\subsubsection{Updating Tracklets Using YOLO Detections}
After a tracklet is instantiated in \Cref{sec:find3d}, we try to update it with every timestep. To map detections to tracklets, we model each YOLO-detection as a line in 3D space and calculate which tracklet of the same class has a belief state closest to it. If the distance is small enough, the EKF of that tracklet performs a measurement step using that detection's pixel coordinates as a measurement. To deal with noise, Kalman Filters use information about the measurement noise $\sigma$, which is assumed to be normally distributed. Since the YOLO network gives us a confidence value in the range $[0,1]$, our EKF can use a modified $\sigma^\prime$, which is defined as $\sigma \cdot (2-yolo\_conf)^{10}$. Therefore, low-confidence detections by the YOLO network have a lower impact on the belief state of the EKF. Note: such a measurement step uses only one detection of one camera. Therefore, we can update an object's position even if we only see it in one camera. (Although the covariance/uncertainty in the direction from that camera to the object will increase.)

After iterating over all detections, all EKFs perform a dynamic step, modeling how the confidence of the belief state decreases with the passing of a timestep. All detections that were not used by an EKF are processed as described in \Cref{sec:find3d}. Also, all tracklets that were not updated recently are removed from the list of active EKF's.

After iterating over all timesteps, we get a list of tracklets that describe the object's positions in 3D space over time.

\section{Data Produced By Our Pipeline}
\label{sec:track3d}
After running the pipeline on the Table Setting Dataset, it's outputs are included in the TSD and will be publicly available with future publications of the Table Setting Dataset. This will include the following files, where \texttt{xxx} and \texttt{yy} are the session and trial id:
\begin{itemize}
    \item \textbf{Video frames of 6 cameras} \\
    \texttt{sxxxtyy.video.<camera\_name>.data.mp4} \\
    \cref{sec:data}: Six \texttt{mp4} files that have been exactly synchronized by using the timestamps recorded with each frame. This is the prior work our pipeline takes as input.
    \item \textbf{YOLO object detections in 2D images} \\
    \texttt{sxxxtyy.mocap.objecttracking.} \texttt{<camera\_name>.yolodetections.yaml} \\
    \cref{sec:yolo}: Six \texttt{yaml} files containing a list. Each element contains the timestamp and a list of detections. Each detection contains the object's name, the \texttt{xywh} of the bounding box and a confidence value. (Although width and height are not used for this pipeline.)
    \item \textbf{Annotations of fixed points in image} \\
    \texttt{sxxxtyy.mocap.objecttracking.} \texttt{imageannotations.yaml} \\
    \cref{sec:annot}: One \texttt{yaml} file containing a nested dictionary, where each camera is a key. For each camera, a dictionary contains lists of \texttt{xy} positions for each annotated point. The points are the 4 corners of each table as well as the world origin.
    \item \textbf{Exact camera poses} \\
    \texttt{sxxxtyy.mocap.objecttracking.} \texttt{cameraposes.yaml} \\
    \cref{sec:optimize_parameters}: One \texttt{yaml} file containing the pose for each camera as well as \texttt{xy}-offsets of the tables relative to their expected position.
    \item \textbf{Trajectories} \\
    \texttt{sxxxtyy.mocap.objecttracking.} \texttt{3dtrajetories.yaml} \\
    \cref{sec:track3d}: One \texttt{yaml} file containing a list of tracklets. Each tracklet contains the object's name as well as a list of timesteps. Each timestep contains the time, \texttt{xyz}-position as well as the covariance matrix of the position as a confidence score. \\
    \texttt{sxxxtyy.mocap.objecttracking.} \texttt{<3dtrajetories|finalpositions>.png} \\
    These are two plots that provide an overview of the trajectories and the object positions at the end of the trial. Examples are shown in \Cref{fig:trajetories,fig:tracked_objects}.
\end{itemize}
\section{Evaluation}
\label{sec:evaluation}
Since the tableware objects we track are not fitted with any sensors, we have no ground truth data to evaluate our pipeline on the TSD. However, we did get access to the TSD lab to record one trial with markers of the motion capture suit attached to tableware objects.

\subsection{Comparison With Optitrack}
\begin{figure}[h]
    \centering
    \begin{subfigure}[b]{1\linewidth}
        \centering
        \includegraphics[clip, trim=0cm 0cm 0cm 0cm, width=1.00\textwidth]{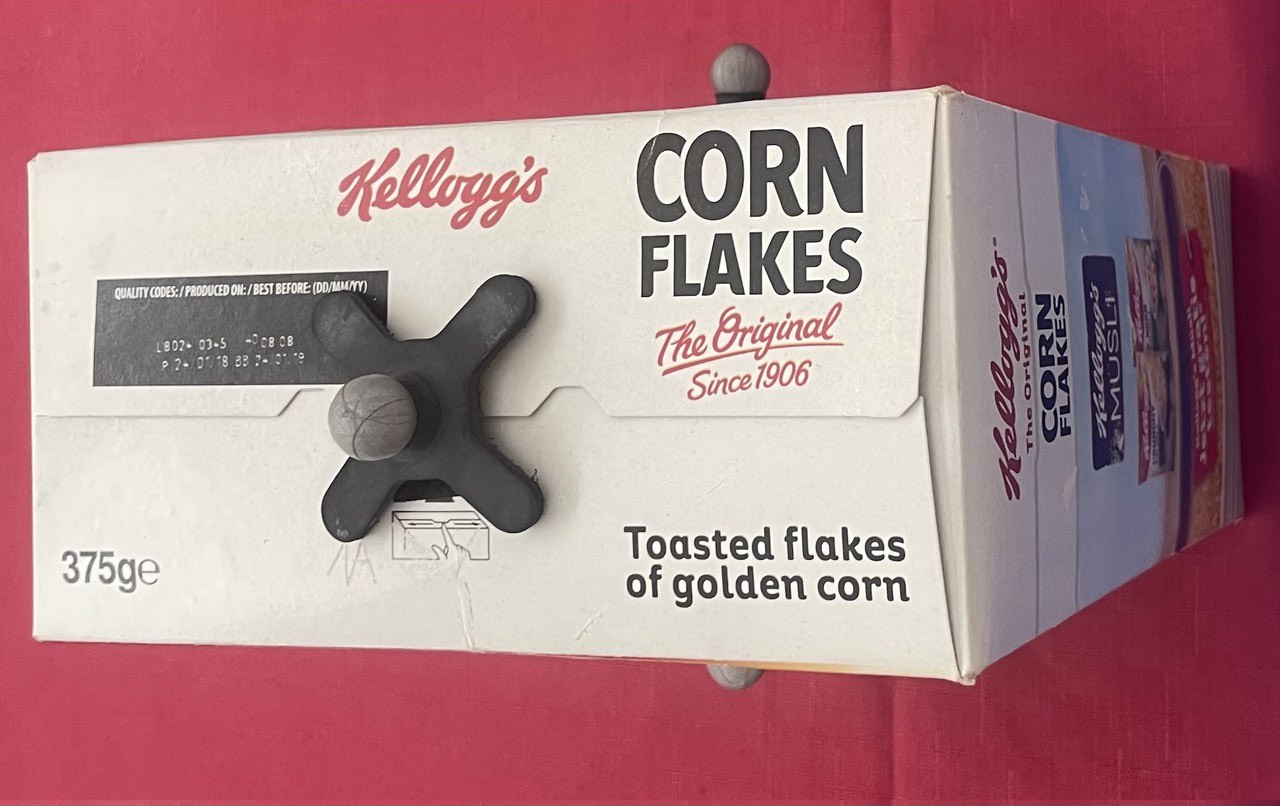}
        \caption{Tableware object with 3 markers for Optitrack system}
        \label{fig:cereal_with_markers}
    \end{subfigure}
    \begin{subfigure}[b]{1\linewidth}
        \centering
        \includegraphics[clip, trim=17cm 12cm 10cm 0cm, width=1.00\textwidth]{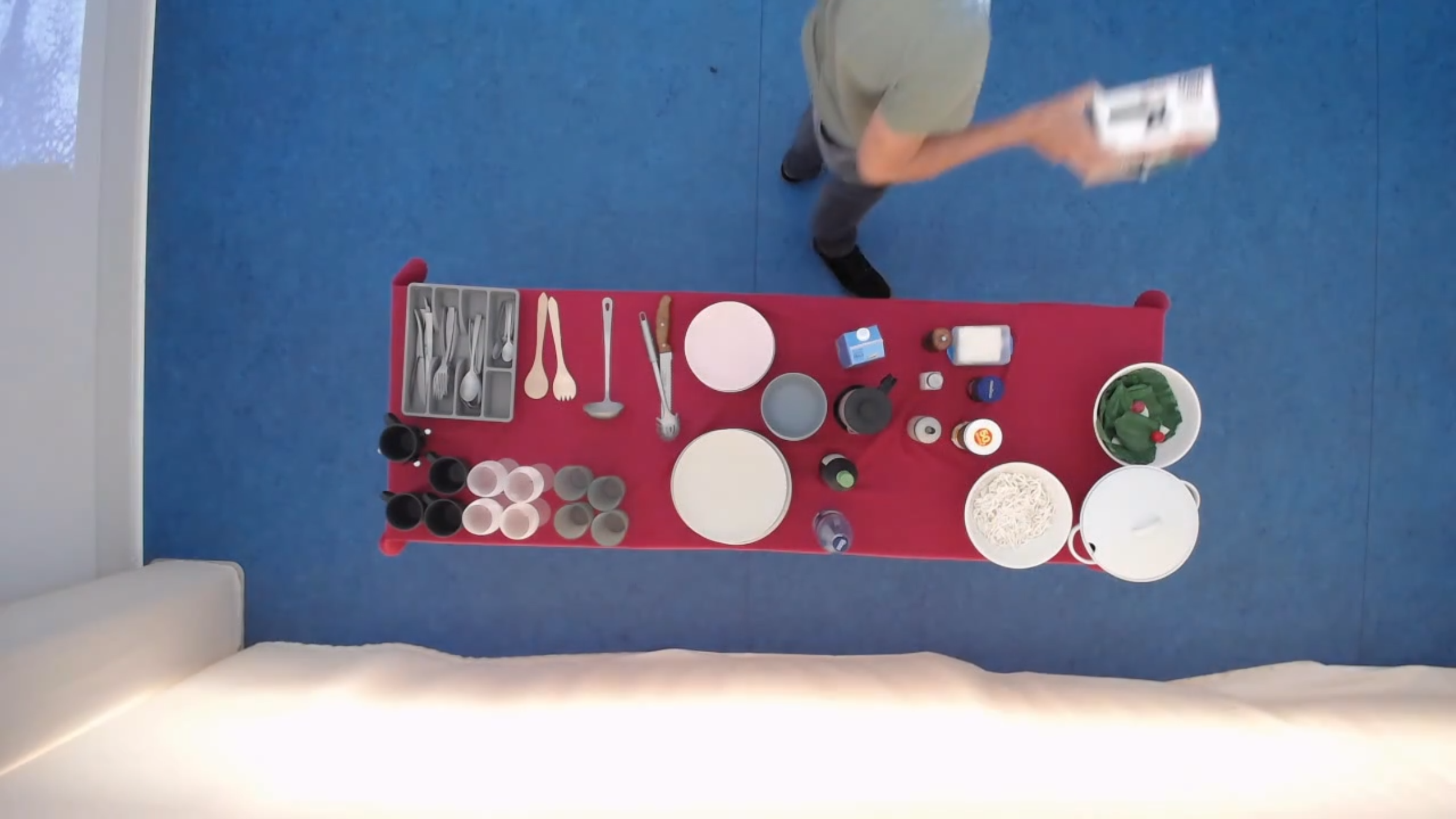}
        \caption{Motion blur effecting the object's visibility}
        \label{fig:cereal_blur}
    \end{subfigure}
    \caption{Motion tracking markers are added to the cereal box}
    \label{fig:validation_setup}
\end{figure}

The TSD uses an Optitrack~\cite{optitrack} motion capture system to track the markers on the test subject's motion capture suit. We got access to the lab for a day and recorded a trial with a special setup for evaluating our pipeline. The motion capture suit was discarded and some of it's markers placed on tableware objects. The bread, cereal and a cup of coffee were fitted with three motion capture markers each, which were modeled as rigid bodies in the recording software Motive~\cite{motive}. However, only the box of cereal was tracked successfully by Optitrack, with the bread and cup of coffee not producing any trajectories. Therefore, we only focus on the box of cereal shown in~\Cref{fig:cereal_with_markers}. As our recording did not follow the official recording process for of the TSD, the ceiling lights where turned off, leading to longer exposure times and therefore increased motion blur visible in~\Cref{fig:cereal_blur}.

\begin{figure}
    \centering
    \begin{subfigure}[b]{1\linewidth}
        \centering
        \includegraphics[clip, trim=0.4cm 0.2cm 1.6cm 1.3cm, width=1.00\textwidth]{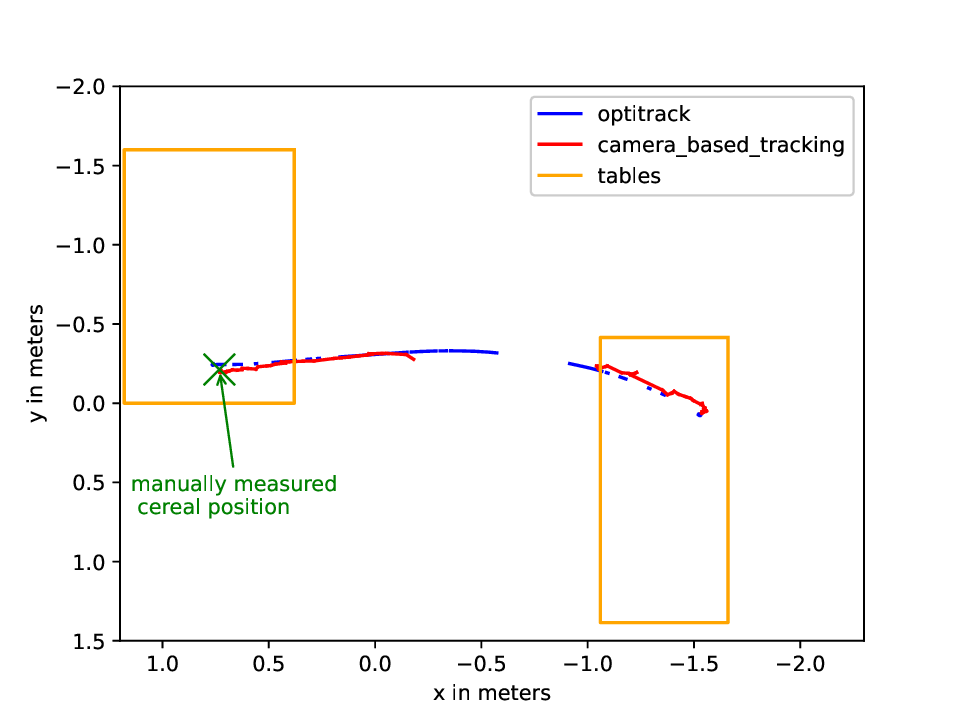}
        \caption{top-down view of Object Trajectory}
        \label{fig:cereal_top_down}
    \end{subfigure}
    \begin{subfigure}[b]{1\linewidth}
        \centering
        \includegraphics[clip, trim=0.4cm 0.2cm 1.6cm 1.3cm, width=1.00\textwidth]{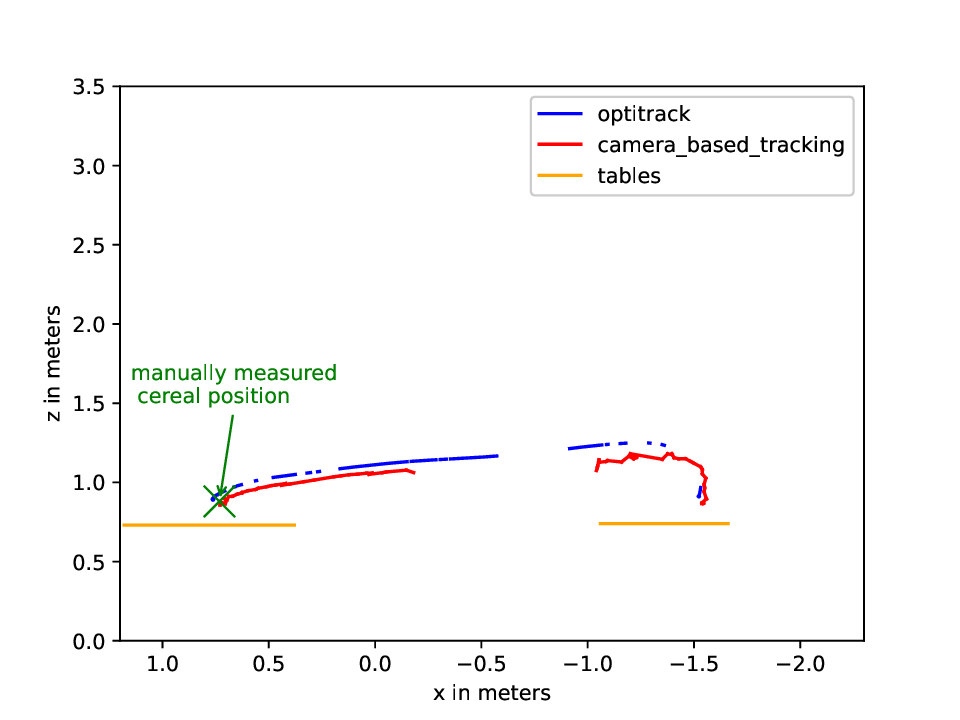}
        \caption{Side view of Object Trajectory}
        \label{fig:cereal_side}
    \end{subfigure}
    \caption{Comparison of marker-based Optitrack system with our rgb-camera-based system}
    \label{fig:cereal}
\end{figure}

Since the Optitrack system and our pipeline use the same world coordinate system, we can easily overlay their trajectories as shown in~\Cref{fig:cereal}. The top-down projection of~\Cref{fig:cereal_top_down} shows that the Optitrack system agrees with our camera based pipeline to within a few centimeters. We also see that both systems suffer from dropouts as the object is moved at around one meter per second. The Optitrack trajectory also drops out many more times at the start and end of the trajectory. This makes calculating a distance between the two trajectories difficult, as we cannot just use the distance of a point on one trajectory to the nearest point of the other trajectory. However, analyzing \Cref{fig:cereal_top_down} gives us confidence in the accuracy of our pipeline.

\Cref{fig:cereal_side} shows the Optitrack trajectory to be about 10 Centimeters above that of our proposed pipeline. We attribute this to the fact that the motion capture markers are all attached to the upper half of the cereal box. Apart from this offset, both tracking systems agree in all three dimensions.

In order to not only rely on Optitrack for our ground truth, we also used a folding rule to measure the final position of the cereal box at the end of the trial. This confirms that both tracking systems are calibrated to the correct coordinate system defined by the tape on the floor at the world coordinate's origin.

These tests show that our pipeline worked well with all tested objects, while the marker-based motion capture system did not track all objects. Such problems would likely get worse with a higher number of objects to be tracked, as the objects with similar markers would not be discriminable from each other.

\subsection{Performance on TSD trials}
\begin{figure*}
    \centering
    \begin{subfigure}[b]{0.36\textwidth}
        \centering
        \includegraphics[clip, trim=12.15cm 0.7cm 9cm 5.6cm, angle=90, width=\textwidth]{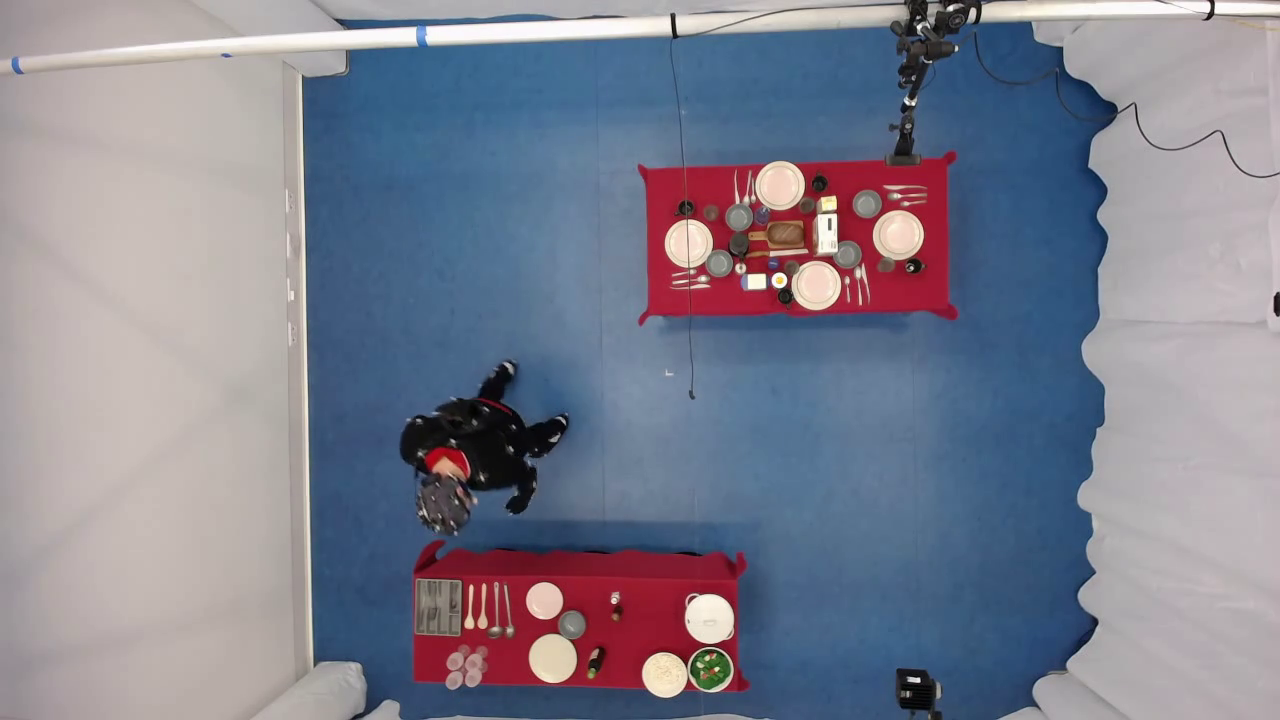}
        \caption{View from Ceiling Camera}
        \label{fig:ceiling_camera}
    \end{subfigure}
    \begin{subfigure}[b]{0.63\textwidth}
        \centering
        \includegraphics[clip, trim=1cm 0.7cm 1.3cm 1.4cm, width=1.00\textwidth]{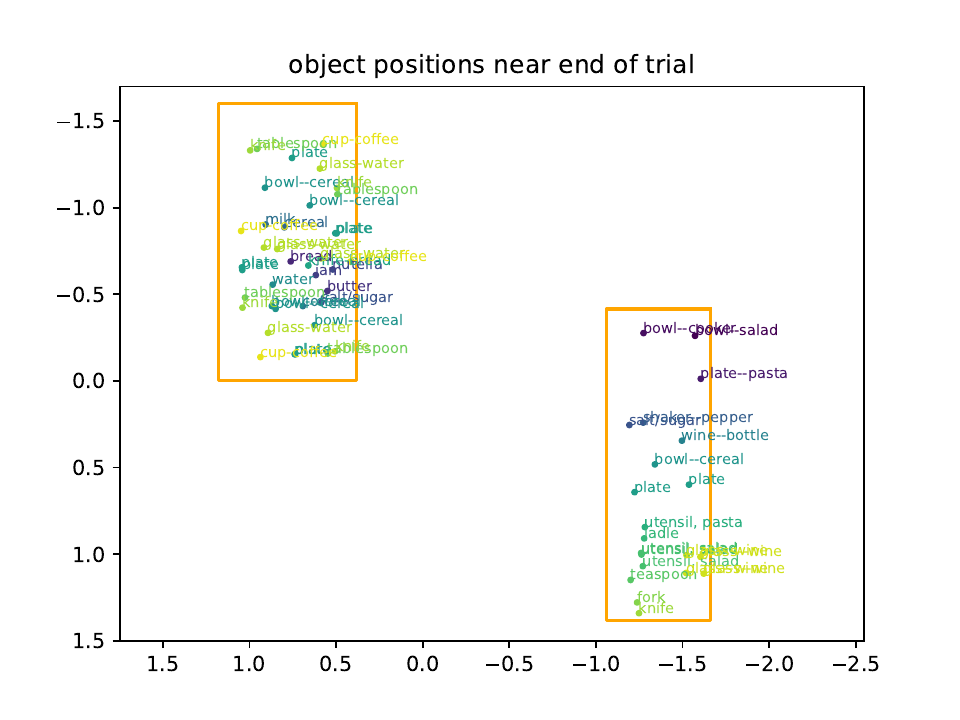}
        \caption{Top-down view of objects tracked by our pipeline}
        \label{fig:tracked_objects}
    \end{subfigure}
    \caption{Comparing camera image with tracked objects for the end of session 28, trial 1}
    \label{fig:finished}
\end{figure*}

In the following, our pipeline will be evaluated on the Table Setting Dataset. Due to the lack of ground truth data regarding the tableware objects, we have to evaluate the trajectories' plausibility manually using images and videos of the webcams. \Cref{fig:finished} shows a side-by-side comparison of the ceiling camera with our pipeline's trajectories just before the end of a trial. We picked a trial with especially many tableware objects set close to each other to demonstrate our pipeline's abilities. With the exception of small teaspoons, all classes of tableware objects are localized accurately. This even works with the wineglasses on the bottom right, that are partially overlapping each other.

\begin{figure}
    \begin{center}
        \includegraphics[clip, trim=3.5cm 1cm 3.5cm 5.25cm, width=1.00\linewidth]{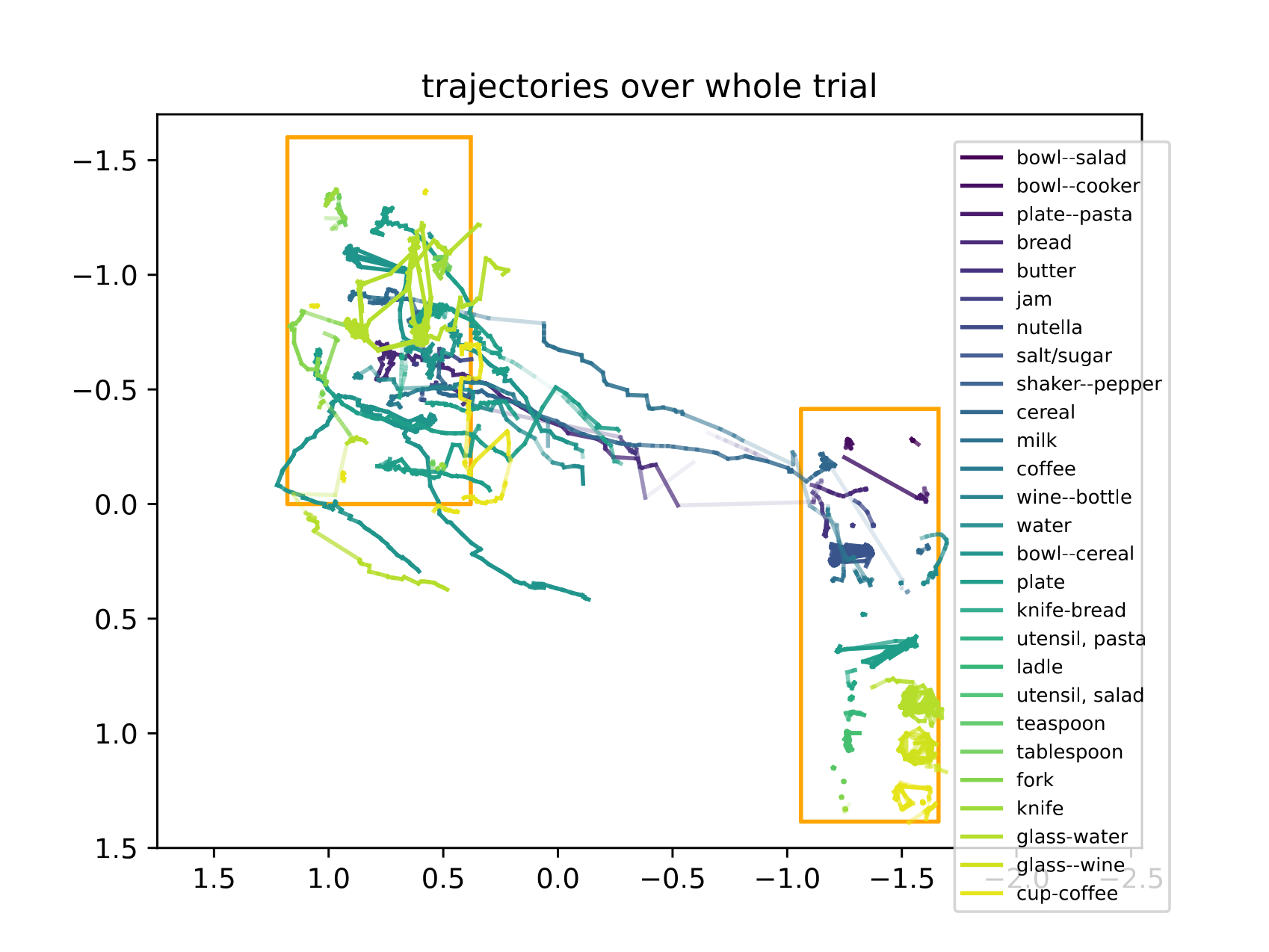}
        \caption{Top-down view of trajectories for session 28, trial 1. High opacity correlates to a high confidence.} \label{fig:trajetories}
    \end{center}
\end{figure}

\Cref{fig:trajetories} shows an example plot with a top-down view of all trajectories of one trial. The confidence of the EKF is visualized using opacity. Here we can see that objects often are not tracked initially after being picked up from the table. However, tracking tends to resume once the objects get close to the table where they are put down. This might be due to the fact that objects are moved at a higher speed initially, causing motion blur. Also, the table being set is better covered by cameras than the space between the tables.

These observations give us confidence, that our trajectories can be used to analyze which objects the test subjects manipulate and where they are placed on the table. The pipeline's performance on trials with missing cameras shows that it is robust to missing cameras. It can be used with any number of cameras as long as the 3-camera constraint of \Cref{sec:find3d} is met.





\section{Future Work}
\label{sec:future_work}
We could include the Point of view (PoV) camera. For our tracking pipeline, we need to know the exact position of the cameras in three-dimensional space. We did not use the PoV camera, since it moves freely in three-dimensional space. But since we do have motion tracking data of the subject's head, we should be able to infer the PoV camera's position for each timestep. We will somehow need to calculate this position in relation to the motion tracking markers of the head, but this should be possible. The challenge lies in the complexity of having a moving camera, where the camera's pose must be extracted from the motion tracking data for each timestep. This is made more difficult by the fact that the soft fabric of the eeg cap can move over time, also moving the markers of the motion capture suit. This would probably require our system to track this slow drifting movement, comparing positions of objects that are tracked both in the stationary and the POV camera. With this ongoing calibration, the POV camera could be used to track objects, that are not trackable otherwise because they are not visible in enough stationary cameras.

\section{Conclusion}
\label{sec:conclusion}
In this paper, we implement a robust, cheap, and scalable pipeline that tracks objects in 3D space, using overlapping, time-synced rgb cameras. A YOLO network is trained and used for detecting objects in the camera's images. Camera position and orientation are calculated for 6 cameras in $320$ trials, using manual annotations of table corners and the world origin marker. Information of these two previous steps is combined to detect objects in 3D space. Tracking objects over time is achieved by implementing an Extended Kalman Filter, which takes YOLO detections as measurements to update object positions over time. This gives us high quality 3D trajectories, with the EKF's covariance being a great confidence metric. In total, $35.416$ tracklets where detected over $320$ trials. The resulting trajectory data will be included in future releases of the Everyday Activity in Science and Engeneering (EASE) Table Setting Dataset (TSD). The Code of the pipeline proposed in this paper is available as supplementary material.
\newpage
{
    \small
    \bibliographystyle{ieeenat_fullname}
    \bibliography{main}
}

\section{Acknowledgements}

This paper has been supported by the German Research Foundation DFG, as part of the Collaborative
Research Center (Sonderforschungsbereich) 1320 EASE - Everyday Activity Science and Engineering, University of Bremen
\url{http://www.ease-crc.org/}. The research was conducted in sub-project H03 Descriptive models of human everyday activity".


\end{document}